\documentclass[10pt,twocolumn,letterpaper]{article}

\usepackage{cvpr}
\usepackage{times}
\usepackage{epsfig}
\usepackage{graphicx}
\usepackage{amsmath}
\usepackage{amssymb}
\usepackage{url}
\usepackage{enumitem}
\usepackage{textcomp}
\usepackage{mathtools}
\usepackage{multirow}
\usepackage{slashbox}
\usepackage{subfig}
\usepackage{dblfloatfix}

\usepackage{caption}
\usepackage[pagebackref=true,breaklinks=true,letterpaper=true,colorlinks,bookmarks=false]{hyperref}

\cvprfinalcopy 

\def\cvprPaperID{1514} 

\ifcvprfinal\fi
\begin{document}

\title{Deep Gaussian Conditional Random Field Network: A Model-based Deep Network for Discriminative Denoising}

\author{Raviteja Vemulapalli\\
University of Maryland, College Park\\
{\tt\small raviteja@umd.edu}
\and
Oncel Tuzel\\
Mitsubishi Electric Research Lab.\\
Cambridge, MA\\
{\tt\small oncel@merl.com}
\and
Ming-Yu Liu\\
Mitsubishi Electric Research Lab.\\
Cambridge, MA\\
{\tt\small mliu@merl.com}
}

\maketitle

\begin{abstract}
We propose a novel deep network architecture for image\\ denoising based on a Gaussian Conditional Random Field (GCRF) model. In contrast to the existing discriminative denoising methods that train a separate model for each noise level, the proposed deep network  explicitly models the input noise variance and hence is capable of handling a range of noise levels. Our deep network, which we refer to as deep GCRF network, consists of two sub-networks: (i) a parameter generation network that generates the pairwise potential parameters based on the noisy input image, and (ii) an inference network whose layers perform the computations involved in an iterative GCRF inference procedure.\ We train the entire deep GCRF network (both parameter generation and inference networks) discriminatively in an  end-to-end fashion by maximizing the peak signal-to-noise ratio measure. Experiments on Berkeley segmentation and PASCALVOC datasets show that the proposed deep GCRF network outperforms state-of-the-art image denoising approaches for several noise levels.
\end{abstract}

\section{Introduction}
In the recent past, deep networks have been successfully used in various image processing and computer vision applications~\cite{Burger12,Girshick14,Taigman14}. Their success can be attributed to several factors such as their ability to represent complex input-output relationships, feed-forward nature of their inference (no need to solve an optimization problem during run time), availability of large training datasets, etc.\ One of the positive aspects of deep networks is that fairly general architectures composed of fully-connected or convolutional layers have been shown to work reasonably well across a wide range of applications.\ However, these general architectures do not use problem domain knowledge which could be very helpful in some of the applications.

For example, in the case of image denoising, it has been recently shown that conventional multilayer perceptrons are not very good at handling multiple levels of input noise~\cite{Burger12}. When a single multilayer perceptron was trained to handle multiple input noise levels (by providing the noise variance as an additional input to the network), it produced inferior results compared to the state-of-the-art BM3D~\cite{Dabov07} approach. In contrast to this, the EPLL framework of~\cite{Zoran11}, which is a model-based approach, has been shown to work well across a wide range of noise levels. These results suggest that we should work towards bringing deep networks and model-based approaches together. Motivated by this, in this work, we propose a novel deep network architecture for denoising based on a Gaussian conditional random field model that explicitly accounts for the input noise level. 
\begin{figure*}[t]
\center
\includegraphics[scale = 0.53]{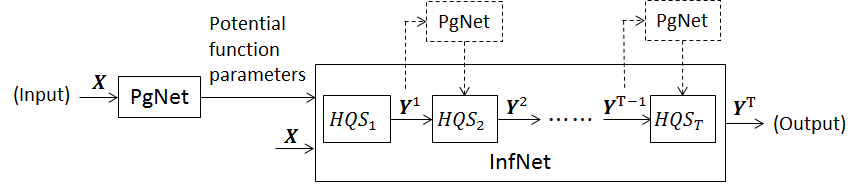}
\caption{The proposed deep GCRF network: Parameter generation network (PgNet) followed by inference network (InfNet). The PgNets in dotted boxes are the additional parameter generation networks introduced after each HQS iteration.}
\label{fig:DGCRF_network}
\end{figure*}

Gaussian Markov Random Fields (GMRFs)~\cite{GMRFbook} are popular models for various structured inference tasks such as denoising, inpainting, super-resolution and depth estimation, as they model continuous quantities and can be efficiently solved using linear algebra routines. However, the performance of a GMRF model depends on the choice of pairwise potential functions. For example, in the case of image denoising, if the potential functions for neighboring pixels are homogeneous (i.e., identical everywhere), then the GMRF model can result in blurred edges and over-smoothed images. Therefore, to improve the performance of a GMRF model, the pairwise potential function parameters should be chosen according to the image being processed. A GMRF model that uses data-dependent potential function parameters is referred to as Gaussian Conditional Random Field (GCRF)~\cite{Tappen07}.

Image denoising using a GCRF model consists of two main steps:\ a \textit{parameter selection step} in which the potential function parameters are chosen based on the noisy input image, and an \textit{inference} \textit{step} in which energy minimization is performed for the chosen parameters. In this work, we propose a novel model-based deep network architecture, which we refer to as \textit{deep GCRF network}, by converting both the parameter selection and inference steps into feed-forward networks.

The proposed deep GCRF network consists of two sub-networks:\ a \textit{parameter generation network (PgNet)} that generates appropriate potential function parameters based on the input image, and an \textit{inference network (InfNet)} that performs energy minimization using the potential function parameters generated by PgNet.\ Since directly generating the potential function parameters for an entire image is very difficult (as the number of pixels could be very large), we construct a full-image pairwise potential function indirectly by combining potential functions defined on image patches. If we use $d \times d$ patches, then our construction defines a graphical model in which each pixel is connected to its $(2d-1) \times (2d-1)$ spatial neighbors.\ This construction is motivated by the recent EPLL framework of~\cite{Zoran11}. Our PgNet directly operates on each $d \times d$ input image patch and chooses appropriate parameters for the corresponding potential function.

Though the energy minimizer can be obtained in closed form for GCRF, it involves solving a linear system with number of variables equal to the number of image pixels (usually of the order of $10^6$). Solving such a large linear system could be computationally prohibitive, especially for\\ dense graphs (each pixel is connected to 224 neighbors when $8\times 8$ image patches are used). Hence, in this work, we use an iterative optimization approach based on \textit{Half Quadratic Splitting} (HQS)~\cite{Geman95, KrishnanF09, Wang08, Zoran11} for designing our inference network. Recently, this approach has been shown to work very well for image restoration tasks even with very few (5-6) iterations~\cite{Zoran11}. Our inference network consists of a new type of layer, which we refer to as HQS layer, that performs the computations involved in a HQS iteration.

Combining the parameter generation and inference networks, we get our deep GCRF network shown in Figure~\ref{fig:DGCRF_network}. Note that using appropriate pairwise potential functions is crucial for the success of GCRF. Since PgNet operates on the noisy input image, it becomes increasingly difficult to generate good potential function parameters as the image noise increases. To address this issue, we introduce an additional PgNet after each HQS iteration as shown in dotted boxes in Figure~\ref{fig:DGCRF_network}. Since we train this deep GCRF network discriminatively in an end-to-end fashion, even if the first PgNet fails to generate good potential function parameters, the later PgNets can learn to generate appropriate parameters based on partially restored images.

\noindent \textbf{Contributions:}
\begin{itemize}[topsep = 2pt, itemsep = 0pt]
\item We propose a new end-to-end trainable deep network architecture for image denoising based on a GCRF model.\ In contrast to the existing discriminative denoising methods that train a separate model for each individual noise level, the proposed network explicitly models the input noise variance and hence is capable of handling a range of noise levels.
\item We propose a differentiable parameter generation network that generates the GCRF pairwise potential parameters based on the noisy input image. 
\item We unroll a half quadratic splitting-based iterative GCRF inference procedure into a deep network and train it jointly with our parameter generation network.
\item We show that, when trained discriminatively by maximizing the peak signal-to-noise (PSNR) measure, the proposed deep network outperforms state-of-the-art image denoising approaches for several noise levels.
\end{itemize}  

\section{Related Work}
\noindent\textbf{Gaussian CRF:} GCRFs were first introduced in~\cite{Tappen07} by modeling the parameters of the conditional distribution of output given input as a function of the input image. The precision matrix associated with each image patch was modeled as a linear combination of twelve derivative filter-based matrices.\ The combination weights were chosen as a parametric function of the responses of the input image to a set of oriented edge and bar filters, and the parameters were learned using discriminative training. This GCRF model was extended to Regression Tree Fields (RTFs) in~\cite{Jancsary12}, where regression trees were used for selecting the parameters of Gaussians defined over image patches.\ These regression trees used responses of the input image to various hand-chosen filters for selecting an appropriate leaf node for each image patch.\ This RTF-based model was trained by iteratively growing the regression trees and optimizing the Gaussian parameters at leaf nodes. Recently, a cascade of RTFs~\cite{SchmidtPAMI15} has also been used for image restoration tasks.\ In contrast to the RTF-based approaches, all the components of our network are differentiable, and hence it can be trained end-­to-­end using standard gradient-based techniques.


Recently,~\cite{SchmidtR14} proposed a cascade of shrinkage fields for image restoration tasks. They learned a separate filter bank and shrinkage function for each stage of their cascade using discriminative training. Though this model can also be seen as a cascade of GCRFs, the filter banks and shrinkage functions used in the cascade do not depend on the noisy input image during test time. In contrast to this, the pairwise potential functions used in our GCRF model are generated by our PgNets based on the noisy input image.

Our work is also related to the EPLL framework of~\cite{Zoran11}, which decomposed the full-image Gaussian model into patch-based Gaussians, and used HQS iterations for GCRF inference.\ Following are the main differences between EPLL and this work: (i) We propose a new deep network architecture which combines HQS iterations with a differentiable parameter generation network.\ (ii)  While EPLL chooses the potential parameters for each image patch as one of the $K$ possible matrices, we construct each potential parameter matrix as a convex combination of $K$ base matrices. (iii) While EPLL learns the $K$ possible potential parameter matrices in a generative fashion by fitting a Gaussian Mixture Model (GMM) to clean image patches, we learn the $K$ base matrices in a discriminative fashion by end-to-end training of our deep network. As shown later in the experiments section, our discriminative model clearly outperforms the generatively trained EPLL.\\[5pt]
\noindent\textbf{Denoising:} Image denoising is one of the oldest problems in image processing and various denoising algorithms have been proposed over the past several years.\ Some of the most popular algorithms include wavelet shrinkage~\cite{simoncelli1996noise}, fields of experts~\cite{RothB09}, Gaussian scale mixtures~\cite{PortillaSWS03}, BM3D~\cite{Dabov07}, non-linear diffusion process-based approaches~\cite{ChenYP15, Hajiaboli11, PlonkaM08}, sparse coding-based approaches~\cite{DongLZS11,DongZSL13,EladA06,Mairal09}, weighted nuclear norm minimization (WNNM)~\cite{GuZZF14}, and non-local Bayesian denoising~\cite{Lebrun13}. Among these, BM3D is currently the most widely-used state-of-the-art denoising approach.\ It is a well-engineered algorithm that combines non-local patch statistics with collaborative filtering.\\[5pt]
\textbf{Denoising with neural networks:} Recently, various deep neural network-based approaches have also been proposed for image denoising~\cite{Agostinelli13, Burger12, JainS08, Xie12, ZhangS05}. While~\cite{JainS08} used a convolutional neural network (CNN),~\cite{Burger12,ZhangS05} used multilayer perceptrons (MLP), and~\cite{Agostinelli13, Xie12} used stacked sparse denoising autoencoders (SSDA).\ \ Among these MLP~\cite{Burger12} has been shown to work very well outperforming the BM3D approach.\ However, none of these deep networks explicitly model the input noise variance, and hence are not good at handling multiple noise levels.\ In all these works, a different network was trained for each noise level.\\[5pt]
\noindent\textbf{Unfolding inference as a deep network:} The proposed approach is also related to a class of algorithms that learn model parameters discriminatively by back-propagating the gradient through a fixed number of inference steps. In~\cite{Barbu09}, the fields of experts~\cite{RothB09} MRF model was discriminatively trained for image denoising by unfolding a fixed number of gradient descent inference steps. In~\cite{RossMHB11}, message-passing inference machines were trained for structured prediction tasks by considering the belief propagation-based inference of a discrete graphical model as a sequence of predictors.\ In~\cite{Gregor10}, a feed-forward sparse code predictor was trained by unfolding a coordinate descent based sparse coding inference algorithm. In~\cite{SchwingU15, Shuai15}, deep CNNs and discrete graphical models were jointly trained by unfolding the discrete mean-field inference. Recently,~\cite{ChenYP15} revisited the classical non-linear diffusion process ~\cite{PeronaM90} by modeling it using several parameterized linear filters and influential functions.\ The parameters of this diffusion process were learned discriminatively by back-propagating the gradient through a fixed number of diffusion process iterations. Though this diffusion process-based approach has been shown to work well for the task of image denoising, it uses a separate model for each noise level, which is undesirable.

In this work, we design our inference network using HQS-based inference of a Gaussian CRF model, resulting in a completely different network architecture compared to the above unfolding works. In addition to this inference network, our deep GCRF network also consists of other sub-networks used for modeling the GCRF pairwise potentials.\\[5pt]
\textbf{Notations:} We use bold face capital letters to denote matrices and bold face small letters to denote vectors. We use vec($\mathbf{A}$), $\mathbf{A}^{\top}$ and $\mathbf{A}^{-1}$ to denote the column vector representation, transpose and inverse of a matrix $\mathbf{A}$, respectively. $\mathbf{A} \succeq 0$ means $\mathbf{A}$ is symmetric and positive semidefinite.
\section{Gaussian Conditional Random Field}
Let $\mathbf{X}$ be the given (noisy) input image and $\mathbf{Y}$ be the (clean) output image that needs to be inferred. Let $\mathbf{X}(i,j)$ and $\mathbf{Y}(i,j)$ represent the pixel $(i,j)$ in images $\mathbf{X}$ and $\mathbf{Y}$, respectively. In this work, we model the conditional probability density $p(\mathbf{Y} | \mathbf{X})$ as a Gaussian distribution given by $p\left(\mathbf{Y} | \mathbf{X}\right)\ \propto \exp\ \left\lbrace - E\left(\mathbf{Y} | \mathbf{X}\right)\right\rbrace$, where
\begin{equation}
\begin{aligned}
E\left(\mathbf{Y} | \mathbf{X}\right) &= \frac{1}{2\sigma^2}\sum_{ij}\left[\mathbf{Y}(i,j) - \mathbf{X}(i,j)\right]^2\hspace{4pt} \Big\} \coloneqq E_d\left(\mathbf{Y} | \mathbf{X}\right)\\
&\hspace{10pt}+ \frac{1}{2}\text{vec}(\mathbf{Y})^{\top} \mathbf{Q(X)}\text{vec}(\mathbf{Y})\hspace{12pt} \Big\} \coloneqq E_p\left(\mathbf{Y} | \mathbf{X}\right).
\end{aligned}
\end{equation}
Here, $\sigma^2$ is the input noise variance and $\mathbf{Q(X)} \succeq 0$ are the (input) data-dependent parameters of the quadratic pairwise potential function $E_p\left(\mathbf{Y} | \mathbf{X}\right)$ defined over the image $\mathbf{Y}$.~\footnote{Note that if the pairwise potential parameters $Q$ are constant, then this model can be interpreted as a generative model with $E_d$ as the data term, $E_p$ as the prior term and $p(\mathbf{Y} | \mathbf{X})$ as the posterior. Hence, our GCRF is a discriminative model inspired by a generative Gaussian model.}

\subsection{Patch-based pairwise potential functions}
Directly choosing the (positive semi-definite) pairwise potential parameters $\mathbf{Q}(\mathbf{X})$ for an entire image $\mathbf{Y}$ is very challenging since the number of pixels in an image could be of the order of $10^6$. Hence, motivated by~\cite{Zoran11}, we construct the (full-image) pairwise potential function $E_{p}$ by combining patch-based pairwise potential functions.

Let $\mathbf{x}_{ij}$ and $\mathbf{y}_{ij}$ be $d^2 \times 1$ column vectors representing the $d \times d$ patches centered on pixel $(i,j)$ in images $\mathbf{X}$ and $\mathbf{Y}$, respectively. Let $\bar{\mathbf{x}}_{ij} = \mathbf{G}\mathbf{x}_{ij}$ and $\bar{\mathbf{y}}_{ij} = \mathbf{G}\mathbf{y}_{ij}$ be the mean-subtracted versions of vectors $\mathbf{x}_{ij}$ and $\mathbf{y}_{ij}$, respectively, where $\mathbf{G} = \mathbf{I} - \frac{1}{d^2}\mathbf{1}\mathbf{1}^{\top}$ is the mean subtraction matrix. Here, $\mathbf{1}$ is the $d^2 \times 1$ vector of ones and $\mathbf{I}$ is the $d^2 \times d^2$ identity matrix. Let
\begin{equation}
V\left(\bar{\mathbf{y}}_{ij}|\bar{\mathbf{x}}_{ij}\right) = \frac{1}{2}\bar{\mathbf{y}}_{ij}^{\top}\left(\mathbf{\Sigma}_{ij}(\bar{\mathbf{x}}_{ij})\right)^{-1} \bar{\mathbf{y}}_{ij},\  \mathbf{\Sigma}_{ij}(\bar{\mathbf{x}}_{ij})\succeq 0,
\end{equation}
be a quadratic pairwise potential function defined on patch $\bar{\mathbf{y}}_{ij}$, with $\mathbf{\Sigma}_{ij}(\bar{\mathbf{x}}_{ij})$ being the corresponding (input) data-dependent parameters.\ Combining the patch-based potential functions at all the pixels, we get the following full-image pairwise potential function:
\begin{equation}
\begin{aligned}
E_{p}\left(\mathbf{Y} | \mathbf{X}\right) &= \sum_{ij} V\left(\bar{\mathbf{y}}_{ij}|\bar{\mathbf{x}}_{ij}\right)\\
&= \frac{1}{2}\sum_{ij} \mathbf{y}_{ij}^{\top}\mathbf{G}^{\top} \left(\mathbf{\Sigma}_{ij}(\bar{\mathbf{x}}_{ij})\right)^{-1}\mathbf{G}\mathbf{y}_{ij}.
\end{aligned}
\label{eqn::full-image-prior}
\end{equation}
Note that since we are using all $d \times d$ image patches, each pixel appears in $d^2$ patches that are centered on its $d \times d$ neighbor pixels. In every patch, each pixel interacts with all the $d^2$ pixels in that patch. This effectively defines a graphical model of neighborhood size $(2d-1) \times (2d-1)$ on image $\mathbf{Y}$.
\subsection{Inference}
\label{sec::GMRFInference}
Given the (input) data-dependent parameters $\lbrace\mathbf{\Sigma}_{ij}(\bar{\mathbf{x}}_{ij})\rbrace$ of the pairwise potential function $E_p\left(\mathbf{Y} | \mathbf{X}\right)$, the Gaussian CRF inference solves the following optimization problem:
\begin{equation}
\begin{aligned}
\mathbf{Y}^* = \underset{\mathbf{Y}}{\text{argmin}}\ \sum_{ij}\left. \begin{cases} &\hspace{-10pt}\frac{1}{\sigma^2}\left[\mathbf{Y}(i,j) - \mathbf{X}(i,j)\right]^2\\
& \hspace{-10pt}+\ \mathbf{y}_{ij}^{\top}\mathbf{G}^{\top}\left(\mathbf{\Sigma}_{ij}(\bar{\mathbf{x}}_{ij})\right)^{-1}\mathbf{G}\mathbf{y}_{ij}\end{cases}\right\rbrace.
\end{aligned}
\label{eqn::GMRF-opt}
\end{equation}
Note that the optimization problem~\eqref{eqn::GMRF-opt} is an unconstrained quadratic program and hence can be solved in closed form.\ However, the closed form solution for $\mathbf{Y}$ requires solving a linear system of equations with number of variables equal to the number of image pixels. Since solving such  linear systems could be computationally prohibitive for large images, in this work, we use a half quadratic splitting-based iterative optimization method, that has been recently used in~\cite{Zoran11} for solving the above optimization problem. This approach allows for efficient optimization by introducing auxiliary variables. 

Let $\mathbf{z}_{ij}$ be an auxiliary variable corresponding to the patch $\mathbf{y}_{ij}$. In half quadratic splitting method, the cost function in~\eqref{eqn::GMRF-opt} is modified to
\begin{equation}
J(\mathbf{Y}, \{\mathbf{z}_{ij}\}, \beta) = \sum_{ij}\left.\begin{cases}&\hspace{-10pt}\frac{1}{\sigma^2}\left[\mathbf{Y}(i,j) - \mathbf{X}(i,j)\right]^2\\ & \hspace{-10pt} +\ \beta\|\mathbf{y}_{ij} - \mathbf{z}_{ij}\|_2^2\\ & \hspace{-10pt}+\ \mathbf{z}_{ij}^{\top}\mathbf{G}^{\top}\left(\mathbf{\Sigma}_{ij}(\bar{\mathbf{x}}_{ij})\right)^{-1} \mathbf{G}\mathbf{z}_{ij} \end{cases}\right\rbrace.
\label{eqn::HQScost}
\end{equation}
Note that as $\beta \rightarrow \infty$, the patches $\{\mathbf{y}_{ij}\}$ are restricted to be equal to the auxiliary variables $\{\mathbf{z}_{ij}\}$, and the solutions of~\eqref{eqn::GMRF-opt} and~\eqref{eqn::HQScost} converge.
\begin{figure*}
\center
\begin{minipage}{\linewidth}
\center
\includegraphics[scale=0.4]{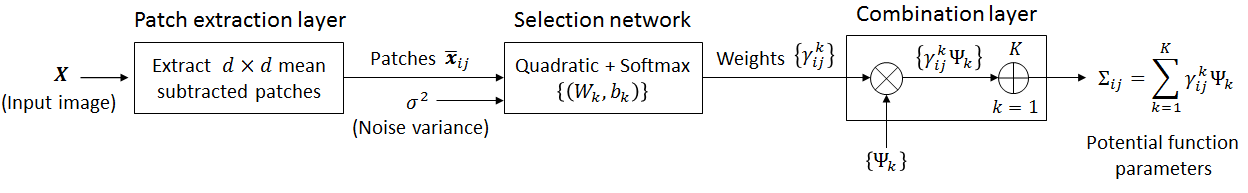}
\vspace{-5pt}
\caption{Parameter generation network: Mean subtracted patches $\bar{\mathbf{x}}_{ij}$ extracted from the input image $\mathbf{X}$ are used to compute the combination weights $\{\gamma_{ij}^k\}$, which are used for generating the pairwise potential parameters $\{\mathbf{\Sigma}_{ij}\}$.}
\label{fig:prior_generation}
\end{minipage}\\[20pt]
\begin{minipage}{\linewidth}
\center
\includegraphics[scale=0.4]{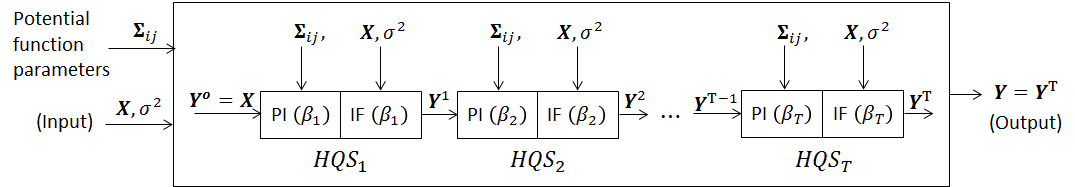}
\caption{Inference network uses the pairwise potential parameters $\{\mathbf{\Sigma}_{ij}(\bar{\mathbf{x}}_{ij})\}$ generated by the PgNet and performs $T$ HQS iterations.}
\label{fig:GCRF_network}
\end{minipage}
\end{figure*}
For a fixed value of $\beta$, the cost function $J$ can be minimized by alternatively optimizing for $\mathbf{Y}$ and $\{\mathbf{z}_{ij}\}$. If we fix $\mathbf{Y}$, then the optimal $\mathbf{z}_{ij}$ is given by 
\begin{equation}
\begin{aligned}
f(\mathbf{y}_{ij}) &= \underset{\mathbf{z}_{ij}}{\text{argmin}}\ \left. \begin{cases} & \hspace{-10pt}\mathbf{z}_{ij}^{\top}\mathbf{G}^{\top}\left(\mathbf{\Sigma}_{ij}(\bar{\mathbf{x}}_{ij})\right)^{-1} \mathbf{G}\mathbf{z}_{ij}\\
&\hspace{-10pt}+\ \beta\|\mathbf{y}_{ij} - \mathbf{z}_{ij}\|_2^2\end{cases}\right\rbrace\\
&= \left(\mathbf{G}^{\top}\left(\mathbf{\Sigma}_{ij}(\bar{\mathbf{x}}_{ij})\right)^{-1} \mathbf{G} + \beta \mathbf{I} \right)^{-1} \beta \mathbf{y}_{ij}\\
& = \left(\mathbf{I} - \mathbf{G}^{\top}\left(\mathbf{\beta\Sigma}_{ij}(\bar{\mathbf{x}}_{ij}) + \mathbf{G}\mathbf{G}^{\top}\right)^{-1}\mathbf{G}\right) \mathbf{y}_{ij}.
\end{aligned}
\label{eqn::optz}
\end{equation}
The last equality in~\eqref{eqn::optz} follows from Woodbury matrix identity. If we fix $\{\mathbf{z}_{ij}\}$, then the optimal $\mathbf{Y}(i,j)$ is given by 
\begin{equation}
\begin{aligned}
g(\{\mathbf{z}_{ij}\}) &= \underset{\mathbf{Y}(i,j)}{\text{argmin}}\ \left. \begin{cases}&\hspace{-10pt} \frac{1}{\sigma^2}\left[\mathbf{Y}(i,j) - \mathbf{X}(i,j)\right]^2\\ & \hspace{-10pt} +\ \beta \sum_{p,q=-\lfloor\frac{d-1}{2}\rfloor}^{\lceil\frac{d-1}{2}\rceil} \left[\mathbf{Y}(i,j) - \mathbf{z}_{pq}(i,j)\right]^2\end{cases}\right\rbrace\\
&= \frac{\mathbf{X}(i,j)}{1 + \beta\sigma^2d^2} + \frac{\beta\sigma^2}{1 + \beta\sigma^2d^2} \sum_{p,q=-\lfloor\frac{d-1}{2}\rfloor}^{\lceil\frac{d-1}{2}\rceil} \mathbf{z}_{pq}(i,j),
\end{aligned}
\label{eqn::optY}
\end{equation}
where $\lfloor\ \ \rfloor, \lceil\ \ \rceil$ are the floor and ceil operators, respectively, and $\mathbf{z}_{pq}(i,j)$ is the intensity value of pixel $(i,j)$ according to the auxiliary patch $\mathbf{z}_{pq}$.

In half quadratic splitting approach, the optimization steps~\eqref{eqn::optz} and~\eqref{eqn::optY} are repeated while increasing the value of $\beta$ in each iteration. This iterative approach has been shown to work well in~\cite{Zoran11} for image restorations tasks even with few (5-6) iterations.
\section{Deep Gaussian CRF network}
\label{sec::Network}
As mentioned earlier, the proposed deep GCRF network consists of the following two components:
\begin{itemize}[topsep = 5pt, leftmargin=13pt]
\item \textbf{Parameter generation network:} This network takes the noisy image $\mathbf{X}$ as input and generates the parameters $\{\mathbf{\Sigma}_{ij}(\bar{\mathbf{x}}_{ij})\}$ of pairwise potential function $E_p\left(\mathbf{Y} | \mathbf{X}\right)$.
\item \textbf{Inference network:} This network performs Gaussian CRF inference using the pairwise potential parameters $\{\mathbf{\Sigma}_{ij}(\bar{\mathbf{x}}_{ij})\}\}$ given by the parameter generation network.
\end{itemize}
\subsection{Parameter generation network}
In this work, we model the pairwise potential parameters $\{\mathbf{\Sigma}_{ij}\}$ as convex combinations of $K$ symmetric positive semidefinite  matrices $\mathbf{\Psi}_1, \hdots, \mathbf{\Psi}_K:$ 
\begin{equation}
\mathbf{\Sigma}_{ij} = \sum_{k} \gamma_{ij}^k \mathbf{\Psi}_k,\ \gamma_{ij}^k \geq 0,\ \sum_k \gamma_{ij}^k = 1.
\end{equation}
The combination weights $\{\gamma_{ij}^k\}$ are computed from the mean-subtracted input image patches $\{\bar{\mathbf{x}}_{ij}\}$ using the following two layer selection network:
\begin{equation}
\begin{aligned}
&\text{Layer 1 - Quadratic layer}:\text{For}\ k = 1,2,\hdots, K\\
&\hspace{30pt}s^{k}_{ij}  = -\frac{1}{2}\bar{\mathbf{x}}_{ij}^{\top} \left(\mathbf{W}_k + \sigma^2 \mathbf{I}\right)^{-1}\bar{\mathbf{x}}_{ij} + b_k.
\end{aligned}
\end{equation}
\begin{equation}
\begin{aligned}
&\text{Layer 2 - Softmax layer}:\text{For}\ k = 1,2,\hdots, K\\
&\hspace{30pt}\gamma_{ij}^k = s^k_{ij} \slash \sum_{p=1}^K s^p_{ij}.
\end{aligned}
\end{equation}

Figure~\ref{fig:prior_generation} shows the overall parameter generation network\\ which includes a patch extraction layer, a selection network and a combination layer. Here, $\{\left(\mathbf{W}_k \succeq 0, \mathbf{\Psi}_k \succeq 0, b_k\right)\}$ are the network parameters, and $\sigma^2$ is the noise variance.

Our choice of the above quadratic selection function is motivated by the following two reasons: (i) Since the selection network operates on mean-subtracted patches, it should be symmetric, i.e., both $\bar{\mathbf{x}}$ and $-\bar{\mathbf{x}}$ should have the same combination weights $\{\gamma^k\}$. To achieve this, we compute each $s^k$ as a quadratic function of $\bar{\mathbf{x}}$. (ii) Since we are computing the combination weights using the noisy image patches, the selection network should be robust to input noise. To achieve this, we include the input noise variance $\sigma^2$ in the computation of $\{s^k\}$. We choose the particular form $\left(\mathbf{W}_k + \sigma^2 \mathbf{I}\right)^{-1}$ because in this case, we can (roughly) interpret the computation of $\{s^k\}$ as evaluating Gaussian log likelihoods. If we interpret $\{\mathbf{W}_k\}$ as covariance matrices associated with clean image patches, then $\{\mathbf{W}_k + \sigma^2 \mathbf{I}\}$ can be interpreted as covariance matrices associated with noisy image patches.
\begin{table*}[t]
\center
\renewcommand\arraystretch{1.25}
\begin{scriptsize}
\begin{tabular}{|c|c|c|c|c|c|c|c|c|c|c|c|}
\hline
Test $\sigma$ & $10$ & $15$ & $20$ & $25$ & $30$ & $35$ & $40$ & $45$ & $50$ & $55$ & $60$\\\hline
ClusteringSR~\cite{DongLZS11} & 33.27 & 30.97 & 29.41 & 28.22 & 27.25 & 26.30 & 25.56 & 24.89 & 24.28 & 23.72 & 23.21\\\hline
EPLL~\cite{Zoran11} & 33.32 & 31.06 & 29.52 & 28.34 & 27.36 & 26.52 & 25.76 & 25.08 & 24.44 & 23.84 & 23.27\\\hline
BM3D~\cite{Dabov07} & 33.38 & 31.09 & 29.53 & 28.36 & 27.42 & 26.64 & 25.92 & 25.19 & 24.63 & 24.11 & 23.62\\\hline     
NL-Bayes~\cite{Lebrun13} & 33.46 & 31.11 & 29.63 & 28.41 & 27.42 & 26.57 & 25.76 & 25.05 & 24.39 & 23.77 & 23.18\\\hline
NCSR~\cite{DongZSL13} & 33.45 & 31.20 & 29.56 & 28.39 & 27.45 & 26.32 & 25.59 & 24.94 & 24.35 & 23.85 & 23.38\\\hline
WNNM~\cite{GuZZF14} & \textbf{33.57} & 31.28 & 29.70 & 28.50 & 27.51 & 26.67 & 25.92 & 25.22 & 24.60 & 24.01 & 23.45\\\hline\hline
CSF~\cite{SchmidtR14} & - & - & - & 28.43 & - & - & - & - & - & - & -\\\hline 
TRD~\cite{ChenYP15} & - & 31.28 & - & 28.56 & - & - & - & - & - & - & -\\\hline
MLP~\cite{Burger12} & 33.43 & - & - & \textbf{28.68} & - & \textbf{27.13} & - & - & \textbf{25.33} & - & -\\\hline\hline
$\text{DGCRF}_5$ & 33.53 & \textbf{31.29} & \textbf{29.76} & 28.58 & \textbf{27.68} & 26.95 & \textbf{26.30} & \textbf{25.73} & 25.23 & \textbf{24.76} & \textbf{24.33}\\\hline
$\text{DGCRF}_8$ & \textbf{33.56} & \textbf{31.35} & \textbf{29.84} & \textbf{28.67} & \textbf{27.80} & \textbf{27.08} & \textbf{26.44} & \textbf{25.88} & \textbf{25.38} & \textbf{24.90} & \textbf{24.45}\\\hline
\end{tabular}
\end{scriptsize}
\caption{Comparison of various denoising approaches on 300 test images.}
\label{table:bsd_pascal_results}
\end{table*}
\subsection{Inference network}
We use the half quadratic splitting method described in Section~\ref{sec::GMRFInference} to create our inference network. Each layer of the inference network, also referred to as a HQS layer, implements one half quadratic splitting iteration. Each HQS layer consists of the following two sub-layers:
\begin{itemize}[leftmargin = 10pt, itemsep=3pt]
\item \textbf{Patch inference layer (PI):} This layer uses the current image estimate $\mathbf{Y}^{t}$ and computes the auxiliary patches $\{\mathbf{z}_{ij}\}$ using $f(\mathbf{y}_{ij})$ given in~\eqref{eqn::optz}.
\item \textbf{Image formation layer (IF):} This layer uses the auxiliary patches $\{\mathbf{z}_{ij}\}$ given by the PI layer and computes the next image estimate $\mathbf{Y}^{t+1}$ using $g(\{\mathbf{z}_{ij}\})$ given in~\eqref{eqn::optY}.
\end{itemize}
Let $\{\beta_1, \beta_2,\hdots, \beta_T\}$ be the $\beta$ schedule for half quadratic splitting. Then, our inference network consists of $T$ HQS layers as shown in Figure~\ref{fig:GCRF_network}. Here, $\mathbf{X}$ is the input image with noise variance $\sigma^2$, and $\{\mathbf{\Sigma}_{ij}(\bar{\mathbf{x}}_{ij})\}$ are the (data-dependent) pairwise potential parameters generated by the PgNet.\\[10pt]
\noindent \textbf{Remark:} Since our inference network implements a fixed number of HQS iterations, its output may not be optimal for~\eqref{eqn::GMRF-opt}. However, since we train our parameter generation and inference networks jointly in a discriminative fashion, the PgNet will learn to generate appropriate pairwise potential parameters such that the output after a fixed number of HQS iterations would be close to the desired output.
\subsection{GCRF network}
Combining the above parameter generation and inference networks, we get our full Gaussian CRF network with parameters $\{\left(\mathbf{W}_k \succeq 0, \mathbf{\Psi}_k \succeq 0, b_k\right)\}$. Note that this GCRF network has various new types of layers that use quadratic functions,  matrix inversions and multiplicative interactions, which are quite different from the computations used in standard deep networks.\\[10pt]
\textbf{Additional PgNets:} Note that using appropriate pairwise potential functions is crucial for the success of GCRF. Since the parameter generation network operates on the noisy input image $\mathbf{X}$, it is very difficult to generate good parameters at high noise levels (even after incorporating the noise variance $\sigma^2$ into the selection network). To overcome this issue, we introduce an additional PgNet after each HQS iteration (shown with dotted boxes in Figure~\ref{fig:DGCRF_network}). The rationale behind adding these additional PgNets is that even if the first PgNet fails to generate good parameters, the later PgNets could generate appropriate parameters using the partially restored images. Our final deep GCRF network consists of $T$ PgNets and $T$ HQS layers as shown in Figure~\ref{fig:DGCRF_network}.\\[10pt]
\textbf{Training:} We train the proposed deep GCRF network end-to-end in a discriminative fashion by maximizing the average PSNR measure.\ We use standard back-propagation to compute the gradient of the network parameters. Please refer to the appendix for relevant derivative formulas. Note that we have a constrained optimization problem here because of the symmetry and positive semi-definiteness constraints on the network parameters $\{\mathbf{W}_k\}$ and $\{\mathbf{\Psi}_k\}$. We convert this constrained problem into an unconstrained one by parametrizing $\mathbf{W}_k$ and $\mathbf{\Psi}_k$ as $\mathbf{W}_k = \mathbf{P}_k\mathbf{P}_k^{\top}, \mathbf{\Psi}_k = \mathbf{R}_k\mathbf{R}_k^{\top}$, where $\mathbf{P}_k$ and $\mathbf{R}_k$ are lower triangular matrices, and use limited memory BFGS~\cite{Liu89} for optimization.
\begin{table*}[t]
\renewcommand\arraystretch{1.3}
\begin{scriptsize}
\begin{minipage}{\textwidth}
\center
\begin{tabular}{|c|c|c|c|c|c|c|c|c|c|c|c|c|c|c|}\hline
Test $\sigma$ & ARF & LLSC & EPLL & opt-MRF & ClusteringSR & NCSR & BM3D & MLP & WNNM & CSF & $\text{RTF}_5$ & TRD & $\text{DGCRF}_8$\\
& ~\cite{Barbu09} & ~\cite{Mairal09} & ~\cite{Zoran11} & ~\cite{Chen13} & ~\cite{DongLZS11} & ~\cite{DongZSL13} & ~\cite{Dabov07} & ~\cite{Burger12} & ~\cite{GuZZF14} & ~\cite{SchmidtR14} & ~\cite{SchmidtPAMI15} & ~\cite{ChenYP15} & \\\hline
15 & 30.70 & 31.27 & 31.19 & 31.18 & 31.08 & 31.19 & 31.08 &   -   & 31.37 & 31.24 & - & \textbf{31.43} &  \textbf{31.43}\\\hline
25 & 28.20 & 28.70 & 28.68 & 28.66 & 28.59 & 28.61 & 28.56 & 28.85 & 28.83 & 28.72 & 28.75 & \textbf{28.95} &  \textbf{28.89}\\\hline
\end{tabular}
\caption{Comparison of various denoising approaches on 68 images (dataset of~\cite{RothB09}) under the unquantized setting.}
\label{table:comparison_on_68_images_unquant}
\end{minipage}\\[10pt]
\begin{minipage}{\textwidth}
\center
\begin{tabular}{|c|c|c|c|c|c|c|c|c|c|c|c|c|c|c|}\hline
Test $\sigma$ & ARF & LLSC & EPLL & opt-MRF & ClusteringSR & NCSR & BM3D & NL-Bayes & MLP & WNNM & CSF & $\text{RTF}_5$ & TRD & $\text{DGCRF}_8$\\
& ~\cite{Barbu09} & ~\cite{Mairal09} & ~\cite{Zoran11} & ~\cite{Chen13} & ~\cite{DongLZS11} & ~\cite{DongZSL13} & ~\cite{Dabov07} & ~\cite{Lebrun13} & ~\cite{Burger12} & ~\cite{GuZZF14} & ~\cite{SchmidtR14} & ~\cite{SchmidtPAMI15} & ~\cite{ChenYP15} & \\\hline
15 & 30.65 & 31.09 & 31.11 & 31.06 & 30.93 & 31.13 & 31.03 & 31.06 & - & 31.20 & - & - & \textbf{31.29} & \textbf{31.36}\\\hline
25 & 28.01 & 28.24 & 28.46 & 28.40 & 28.26 & 28.41 & 28.38 & 28.43 & \textbf{28.77} &  28.48 & 28.53  & \textbf{28.74} & 28.63 & 28.73\\\hline
\end{tabular}
\caption{Comparison of various denoising approaches on 68 images (dataset of~\cite{RothB09}) under the quantized setting.}
\label{table:comparison_on_68_images_quant}
\end{minipage}
\end{scriptsize}
\end{table*}
\section{Experiments}
In this section, we use the proposed deep GCRF network\\ for image denoising. We trained our network using a dataset of 400 images (200 images from BSD300~\cite{MartinFTM01} training set and 200 images from PASCALVOC 2012~\cite{EveringhamEGWWZ15} dataset), and evaluated it using a dataset of 300 images (all 100 images from BSD300~\cite{MartinFTM01} test set and 200 additional images from PASCALVOC 2012~\cite{EveringhamEGWWZ15} dataset).\ For our experiments, we used white Gaussian noise of various standard deviations. For realistic evaluation, all the images were quantized to [0-255] range after adding the noise. We use the standard PSNR measure for quantitative evaluation.

We performed experiments with two patch sizes ($5\times 5$ and $8 \times 8$), and the number of matrices $\mathbf{\Psi}_k$ was chosen as 200. Following~\cite{Zoran11}, we used six HQS iterations with $\beta$ values given by $\frac{1}{\sigma^2}[1, 4, 8, 16, 32, 64]$~\footnote{Optimizing the $\beta$ values using a validation set may further improve our performance.}. To avoid overfitting, we regularized the network, by sharing the parameters $\{\mathbf{W}_k, \mathbf{\Psi}_k\}$ across all PgNets. We initialized the network parameters using the parameters of a GMM learned on clean image patches.

We trained two deep GCRF networks, one for low input\\ noise levels ($\sigma \leq 25$) and one for high input noise levels ($25 < \sigma < 60$). For training the low noise network, we used $\sigma = [8, 13, 18, 25]$ and for training the high noise network, we used $\sigma = [30, 35, 40, 50]$. Note that both the networks were trained to handle a range of input noise levels. For testing, we varied the $\sigma$ from 10 to 60 in intervals of 5. 

Table~\ref{table:bsd_pascal_results} compares the proposed deep GCRF network with various state-of-the-art image denoising approaches on 300 test images. Here, $\text{DGCRF}_5$ and $\text{DGCRF}_8$ refer to the deep GCRF networks that use $5\times 5$ and $8\times 8$ patches, respectively.\ For each noise level, the top two PSNR values are shown in boldface style.\ Note that the CSF~\cite{SchmidtR14}, TRD~\cite{ChenYP15} and MLP~\cite{Burger12} approaches train a different model for each noise level. Hence, for these approaches, we report the results only for those noise levels for which the corresponding authors have reported their results.\ As we can see, the proposed deep GCRF network clearly outperforms the ClusteringSR~\cite{DongLZS11}, EPLL~\cite{Zoran11}, BM3D~\cite{Dabov07}, NL-Bayes~\cite{Lebrun13}, NCSR~\cite{DongZSL13}, CSF and TRD approaches on all noise levels, and the WNNM~\cite{GuZZF14} approach on all noise levels except $\sigma = 10$ (where it performs equally well). Specifically, it produces significant improvement in the PSNR compared to the ClusteringSR (0.29 - 1.24 dB), EPLL (0.24 - 1.18 dB), BM3D (0.18 - 0.83 dB), NL-Bayes (0.10 - 1.27 dB), NCSR (0.11 - 1.07 dB) and WNNM (upto 1.0 dB) approaches. Compared to the recent diffusion process-based TRD~\cite{ChenYP15} approach, we improve the PSNR by 0.07 dB for $\sigma = 15$ and 0.11 dB for $\sigma = 25$. Note that while the TRD approach trained separate models for $\sigma = 15$ and $\sigma = 25$, we use the same deep network for both noise levels. The CSF approach of~\cite{SchmidtR14}, which also uses GCRFs, performs poorly (0.24 dB for $\sigma = 25$) compared to our deep network. 

When compared with MLP\cite{Burger12}, which is the state-of-the-art deep networks-based denoising approach, we perform better for $\sigma = [10, 50]$, worse for $\sigma = 35$, and equally well for $\sigma = 25$. However, note that while~\cite{Burger12} uses a different MLP for each specific noise level, we trained only two networks, each of which can handle a range of noise levels. In fact, our single low noise network is able to outperform the MLP trained for $\sigma = 10$ and perform as good as the MLP trained for $\sigma = 25$. This ability to handle a range of noise levels is one of the major benefits of the proposed deep network. Note that though we did not use the noise levels $\sigma = 10, 15, 20, 45$ during training, our networks achieve state-of-the-art results for these $\sigma$. This shows that our networks are able to handle a range of noise levels rather than just fitting to the training $\sigma$. Also, our high noise network performs very well for $\sigma = 55$ and $60$ even though these values are out of its training range. This shows that the proposed \emph{model-based} deep network can also generalize reasonably well for out-of-range noise levels.

\begin{figure}
\center
\includegraphics[scale=0.28]{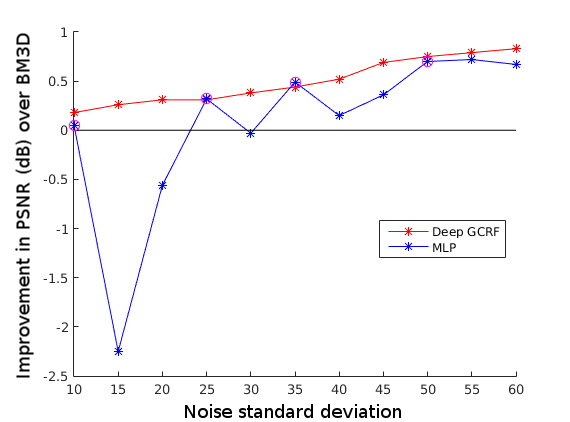}
\caption{Sensitivity analysis of the MLP and the proposed approach. The noise levels for which MLP was trained are indicated using a circular marker.}
\label{fig:mlp_gcrf_comp}
\end{figure}

To analyze the sensitivity of the \emph{non-model} based MLP approach to the deviation from training noise, we evaluated it on noise levels that are slightly ($\pm 5$) different from the training $\sigma$. The authors of~\cite{Burger12} trained separate MLPs for $\sigma = 10, 25, 35, 50$ and $65$.\ As reported in~\cite{Burger12}, training a single MLP to handle multiple noise levels gave inferior results. Figure~\ref{fig:mlp_gcrf_comp} shows the improvement of the MLP approach over BM3D in terms of PSNR. For each noise level, we used the best performing model among $\sigma = 10, 25, 35, 50,65$. As we can see, while the MLP approach does very well for the exact noise levels for which it was trained, it performs poorly if the test $\sigma$ deviates from the training $\sigma$ even by 5 units. This is a major limitation of the MLP approach since training a separate model for each individual noise level is not practical.
In contrast to this, the proposed approach is able to cover a wide range of noise levels just using two networks.

Please note that the purpose of Figure~\ref{fig:mlp_gcrf_comp} is not to compare the performance of our approach with MLP on noise levels that were not used in MLP training, which would be an unfair comparison. The only purpose of this figure is to show that, although very powerful, a network trained for a specific noise level is very sensitive.
\begin{figure*}[!tb]
  \centering
    \begin{tabular}{cccccc}
    \includegraphics[width=.16\textwidth]{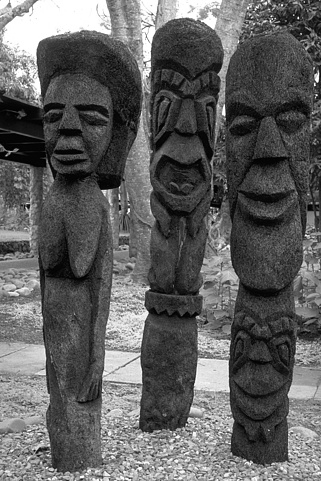} \hspace{-0.45cm} &
    \includegraphics[width=.16\textwidth]{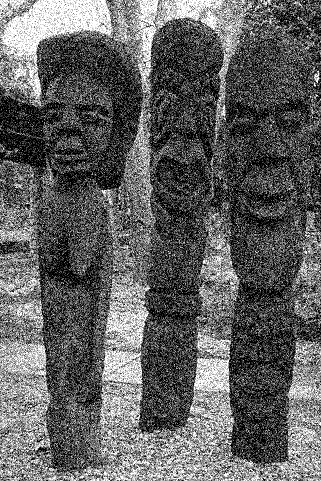} \hspace{-0.45cm} &
    \includegraphics[width=.16\textwidth]{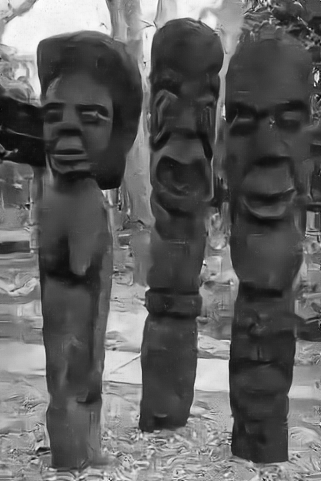} \hspace{-0.45cm} &
    \includegraphics[width=.16\textwidth]{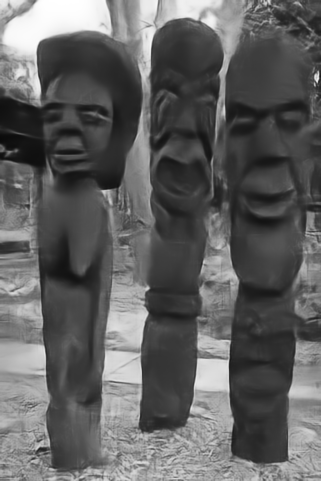} \hspace{-0.45cm} &
    \includegraphics[width=.16\textwidth]{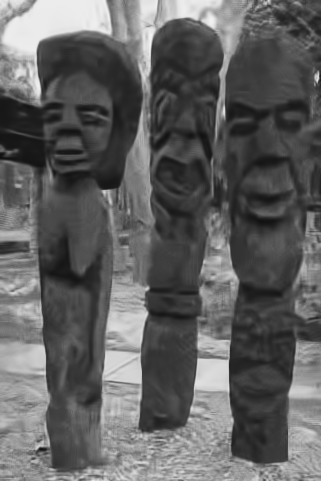} \hspace{-0.45cm} &
    \includegraphics[width=.16\textwidth]{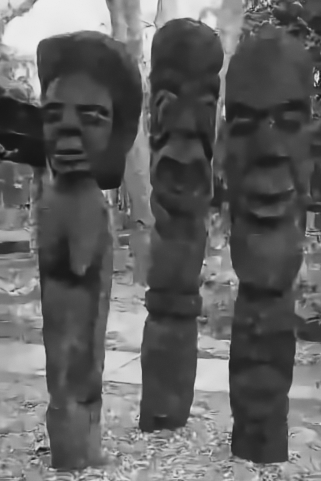} \\
     Original & Noisy & Deep GCRF & MLP & BM3D & EPLL
    \end{tabular}
  \caption{Visual comparison of the proposed approach with benchmark methods for Gaussian noise of $\sigma = 50$.}
\label{fig:images}
\end{figure*} 

Apart from our test set of 300 images, we also evaluated\\ our network on a smaller dataset of 68 images~\cite{RothB09} which has been used in various existing works. Tables~\ref{table:comparison_on_68_images_unquant} and~\ref{table:comparison_on_68_images_quant} compare the proposed deep GCRF network with various approaches on this dataset under the unquantized and quantized settings, respectively. For each noise level, the top two PSNR values are shown in boldface style. As we can see, the proposed approach outperforms all the other approaches except $\text{RTF}_5$~\cite{SchmidtPAMI15} and MLP~\cite{Burger12} under the quantized setting, and TRD~\cite{ChenYP15} under the unquantized setting. However, note that while we use a single network for both $\sigma = 15$ and $\sigma = 25$, the MLP, TRD and $\text{RTF}_5$ approaches trained their models specifically for individual noise levels.

Figure~\ref{fig:images} presents a visual comparison of the proposed approach with various benchmark methods using a test image corrupted with Gaussian noise of $\sigma = 50$. As we can see, the proposed deep GCRF network produces a sharper image with less artifacts compared to other approaches.\\[8pt]
\textbf{Computational time:} For a $321 \times 481$ image, the proposed\\ deep network ($\text{DGRF}_8$) takes 4.4s on an NVIDIA Titan GPU using a MATLAB implementation.
\section{Conclusions}
In this work, we proposed a new end-to-end trainable deep network architecture for image denoising based on a Gaussian CRF model.\ The proposed network consists of a parameter generation network that generates appropriate potential function parameters based on the input image, and an inference network that performs approximate Gaussian CRF inference.\ Unlike the existing discriminative denoising approaches that train a separate model for each individual noise level, the proposed network can handle a range of noise levels as it explicitly models the input noise variance. When trained discriminatively by maximizing the PSNR measure, this network outperformed various state-of-the-art denoising approaches. In the future, we plan to use this network for other full-image inference tasks like super-resolution, depth estimation, etc.
\section*{Appendix}
In this appendix, we show how to back-propagate the loss derivatives through the layers of our deep GCRF network. Let L be the final loss function.\\[10pt]
\textbf{Backpropagation through the combination layer:} Given the derivatives $dL/d\mathbf{\Sigma}_{ij}$ of the loss function $L$ with respect to the pairwise potential parameters $\mathbf{\Sigma}_{ij}$, we can compute the derivatives of $L$ with respect to the combination weights $\gamma_{ij}^k$ and the matrices $\mathbf{\Psi}_k$ using
\begin{equation}
\frac{dL}{d\gamma_{ij}^k} = \text{trace}(\mathbf{\Psi}_k^{\top}\frac{dL}{d\mathbf{\Sigma}_{ij}}),\ \ \frac{dL}{d\mathbf{\Psi}_k} = \sum_{ij} \gamma_{ij}^k \frac{dL}{d\mathbf{\Sigma}_{ij}}.\\
\end{equation}
\textbf{Backpropagation through the patch inference layer:} Given the derivatives $dL/d\mathbf{z}_{ij}$ of the loss function $L$ with respect to the output of a patch inference layer, we can compute the derivatives of $L$ with respect to its input patches $\mathbf{y}_{ij}$ and the pairwise potential parameters $\mathbf{\Sigma}_{ij}$ using
\begin{equation}
\begin{aligned}
&\frac{dL}{d\mathbf{y}_{ij}} = \left(\mathbf{I} - \mathbf{G}(\beta \mathbf{\Sigma}_{ij} + \mathbf{G})^{-1}\mathbf{G}\right)\frac{dL}{d\mathbf{z}_{ij}},\\
&\frac{dL}{d\mathbf{\Sigma}_{ij}} = \beta \left(\beta \mathbf{\Sigma}_{ij} + \mathbf{G} \right)^{-1}\mathbf{G} \frac{dL}{d\mathbf{z}_{ij}} \mathbf{y}_{ij}^{\top} \left(\beta\mathbf{\Sigma}_{ij} + \mathbf{G} \right)^{-1}\mathbf{G}.\\
\end{aligned}
\end{equation}
\noindent\textbf{Backpropagation through the selection network:} Given the derivatives $dL/d\gamma_{ij}^k$ of the loss function $L$ with respect to the combination weights $\gamma_{ij}^k$, we can compute the derivatives of $L$ with respect to the selection network parameters $\left(\mathbf{W}_k, b_k\right)$ and the input patches $\bar{\mathbf{x}}_{ij}$ using:
\begin{equation}
\begin{aligned}
&\frac{dL}{d\mathbf{W}_k} = \left(\mathbf{W}_k + \sigma^2 \mathbf{I}\right)^{-1}\left(\sum_{ij} \frac{dL}{ds_{ij}^k} \frac{\mathbf{x}_{ij}\mathbf{x}_{ij}^{\top}}{2}\right)\left(\mathbf{W}_k + \sigma^2 \mathbf{I}\right)^{-1}\\
&\frac{dL}{db_k} = \sum_{ij} \frac{dL}{ds_{ij}^k},\ \ \ \ \frac{dL}{ds_{ij}^k} = -\gamma_{ij}^k\left(\sum_{\substack{n=1\\n\neq k}}^K \gamma_{ij}^n\frac{dL}{d\gamma_{ij}^n}\right),\\
& \frac{dL}{d\bar{\mathbf{x}}_{ij}} = - \sum_{k=1}^K \frac{dL}{d\gamma_{ij}^k} \left(\mathbf{W}_k + \sigma^2 \mathbf{I} \right)^{-1} \bar{\mathbf{x}}_{ij}.
\end{aligned}
\end{equation}

We skip the derivative formulas for other computations such as softmax, extracting mean-subtracted patches from an image, averaging in the image formation layer, etc., as they are standard operations.

{\small
\bibliographystyle{ieee}
\bibliography{DenoisingArxiv}
}

\end{document}